%% file: main.tex
\documentclass[letterpaper, 10 pt, conference]{ieeeconf}  
\usepackage{newtxtext}

\IEEEoverridecommandlockouts                              

\usepackage{graphics}
\usepackage{multirow}

\usepackage{graphics} 
\usepackage{epsfig} 
\usepackage{times} 
\usepackage{amsmath} 
\usepackage{amssymb}  

\usepackage{array}
\usepackage{booktabs}

\usepackage[hidelinks]{hyperref}
\usepackage{cite}

\usepackage{xcolor,pifont}
\usepackage{color}
\usepackage[nolist]{acronym}
\usepackage{amsmath}
\usepackage{multirow}
\usepackage{microtype}

\newcommand{\figref}[1]{Fig.~\ref{#1}}    

\newcommand{\tabref}[1]{Table~\ref{#1}}

\title{\LARGE \bf
DFM: Deep Fourier Mimic for Expressive Dance Motion Learning
}

\title{\LARGE \bf
FTACT: Force Torque aware Action Chunking Transformer \\ for Pick-and-Reorient Bottle Task
}

\author{Ryo Watanabe$^{1}$, Maxime Alvarez$^{1}$$^{,}$$^{2}$, Pablo Ferreiro$^{1}$, Pavel Savkin$^{1}$, Genki Sano$^{1}$ 
\\
\thanks{$^{1}$ TELEXISTENCE Inc, Foundation Model Division, Japan.}
\thanks{$^{2}$ The University of Tokyo, Japan.}
}

\begin{document}

\maketitle
\thispagestyle{empty}
\pagestyle{empty}

\begin{abstract}
Manipulator robots are increasingly being deployed in retail environments, yet contact–rich edge cases still trigger costly human teleoperation. A prominent example is upright lying beverage bottles, where purely visual cues are often insufficient to resolve subtle contact events required for precise manipulation. We present a multimodal \ac{il} policy that augments the \ac{act} with force and torque sensing, enabling end-to-end learning over images, joint states, and forces and torques. Deployed on \textit{Ghost}, single-arm platform by Telexistence Inc, our approach improves \ac{pnr} bottle task by detecting and exploiting contact transitions during pressing and placement. Hardware experiments demonstrate greater task success compared to baseline matching the observation space of ACT  as an ablation and experiments indicate that force and torque signals are beneficial in the press and place phases where visual observability is limited, supporting the use of interaction forces as a complementary modality for contact–rich skills. The results suggest a practical path to scaling retail manipulation by combining modern \ac{il} architectures with lightweight force and torque sensing.
\end{abstract}

\input{sections/1-intro}

\input{sections/2-system}

\input{sections/3-result}

\input{sections/4-conclusion}
\input{acronyms}

\bibliographystyle{IEEEtran}
\bibliography{main} 

\end{document}

%% file: sections/1-intro.tex
\section{INTRODUCTION}
\begin{figure}[!t]
    \centering
    \includegraphics[width=\linewidth]{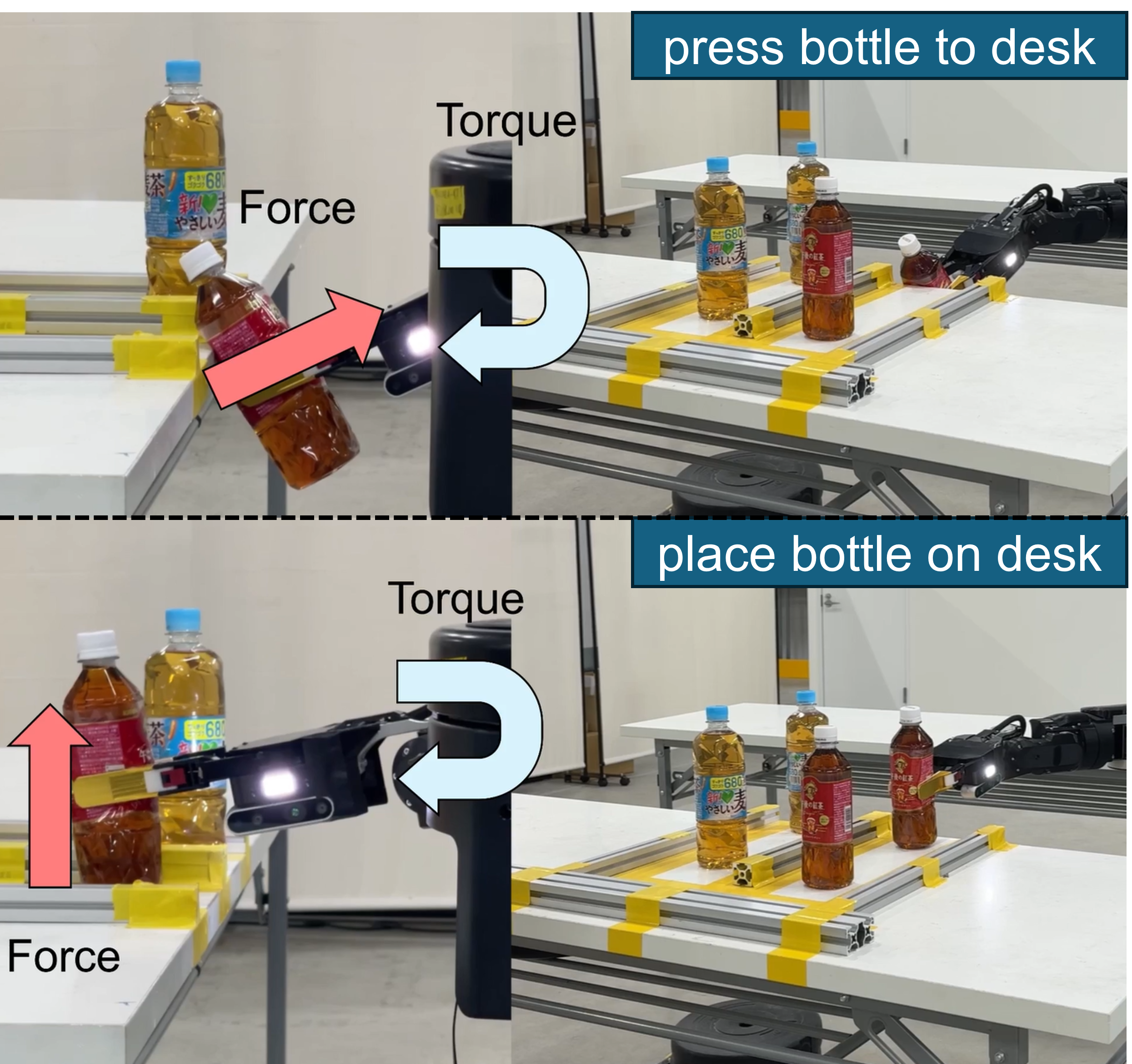}
    \caption{Pick-and-Reorient (PnR) task consists of a contact-rich task such as pressing the bottle to the desk and placing the bottle upright where the force and torque information are helpful in addition to the visual feedback. Youtube link for digest video of our work: \url{https://youtu.be/dKnjDpK1_YU}}
    \label{fig:pick_and_reorient_concept}
    \vspace{-2ex}
\end{figure}

Robotic manipulation is advancing rapidly due to progress in actuators, sensors, and artificial intelligence. These advances enables robots in retail domains such as supermarkets and convenience stores to automate repetitive tasks formerly performed by humans~\cite{ghost, simbe, keon}. For example, Ghost~\cite{ghost}, a robotic manipulator deployed by Telexistence Inc, performs \ac{pnp} of beverage bottles in operating convenience stores. Although much of the workflow is automated by rule-based controllers, certain failure modes remain difficult to recover autonomously. A particularly challenging case occurs when bottles lie on the beverage shelf, with widely varying scene configurations such as different bottle poses and types. In this setting, a single-arm manipulator needs to pick up, reorient, and place the bottle upright—a contact-rich sequence we refer to as \ac{pnr}. In particular, pressing the bottle against the shelf to reorient the bottle and releasing the gripper to determine that the bottle stands upright are difficult to resolve from vision alone, as illustrated in \figref{fig:pick_and_reorient_concept}. As a result, teleoperators using VR headsets occasionally intervene. Even infrequent handovers incur operational costs and hinder scaling the robotic manipulator deployment to a large numbers of stores.

\ac{il} has been studied extensively~\cite{argall2009survey_il, ross2011reduction_il, pomerleau1989alvinn_il}, and recent image-conditioned policies and \ac{vla} models have markedly improved fine-grained manipulation performance~\cite{act,pi0,gr00t,buamanee2024_biact}. However, purely visual and proprioceptive observations can be insufficient for contact-rich skills. Force and torque feedback is especially helpful when visual cues are ambiguous. Prior work has begun to integrate tactile or force sensing into \ac{il} pipelines~\cite{vla-touch,nakahara2025learning,li2025adaptive}, with much of the focus on tactile arrays for dexterous hands. By contrast, the effectiveness of widely used single-arm, gripper-based robots that leverage wrist force and torque sensing remains underexplored. Although some studies incorporate wrist force and torque information into \ac{il}~\cite{liu2024forcemimic,xue2025forcereactive}, they often rely on diffusion policies\cite{chi2023diffusion} with higher computational cost and target household cooking tasks, which differ from \ac{pnr} that comprises the pressing and placement subtasks required for bottle handling in retail.

\begin{figure*}
    \centering
    \vspace{10pt}
    \includegraphics[width=\linewidth]{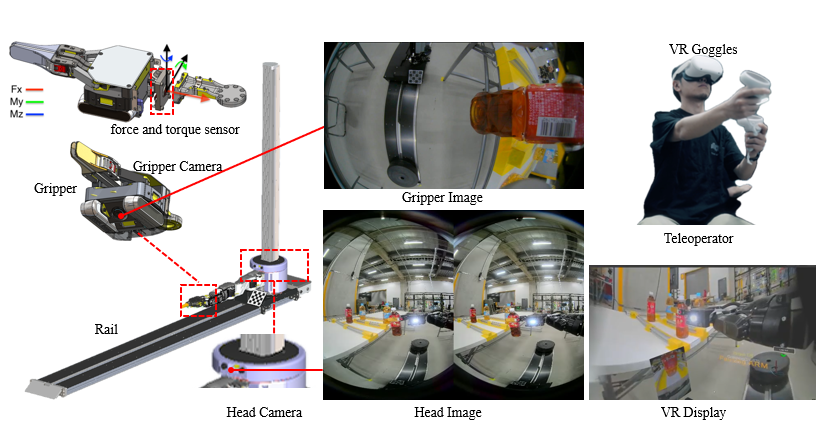}
    \caption{System overview. Left: Hardware of the single-arm with gripper robot, featuring a wrist-mounted force and torque sensor and two camera modules: a downward-tilted gripper camera and a pair of fisheye head cameras. Middle-top: Example image from the gripper camera used for grasping. Middle-bottom: Example stereo fisheye views from the head cameras for global scene awareness. Right-top: Teleoperator wearing a VR headset and hand controllers to collect the dataset for training. Right-bottom: VR display used during teleoperation, rendering the head-camera view with the gripper-camera view overlaid.}
    \label{fig:system_overview}
    \vspace{-2ex}
\end{figure*}

In this work, we augment the Action Chunking Transformer (\ac{act})—a comparatively lightweight alternative to diffusion policies and large \ac{vla} models—with wrist force and torque signals, yielding \ac{ftact}, a multimodal end-to-end policy for contact-rich retail tasks. We validate our approach on Ghost, single-arm gripper-equipped platform, through real-world experiments. Our results indicate that incorporating wrist force and torque sensing improves recovery in fallen bottle scenarios, increasing success rates and reducing the need for teleoperator interventions.

Our main contributions are:
\begin{itemize}
\item An \ac{il} framework that augments \ac{act} with wrist force and torque sensing into a single multimodal policy.
\item A real-world evaluation on a single-arm with gripper retail robot demonstrating superior performance on the fine-grained task of \ac{pnr} such as pressing the bottle and uprighting bottles, along with an analysis of force and torque signals.
\end{itemize}

%% file: sections/2-system.tex
\section{System}
\subsection{Hardware}
The robot has a total of 10 \acp{dof}: seven revolute joints and three prismatic joints mounted on a rail and gripper. Three fisheye cameras are attached to the robot. One camera is mounted below the gripper and tilted by $45^\circ$ to view the bottle during grasping. Two cameras are mounted on the head and can observe both the gripper and the bottle, as shown in \figref{fig:system_overview}. A strain-gauge, force and torque sensor~\cite{howe1992dynamic, brown1983strain} is attached at the base of the gripper as seen in \figref{fig:system_overview}; it measures the wrench such as forces and torques. There are encoders for each joint which can measure the absolute joint state.

\begin{figure}[t]
    \centering
    \includegraphics[width=\linewidth]{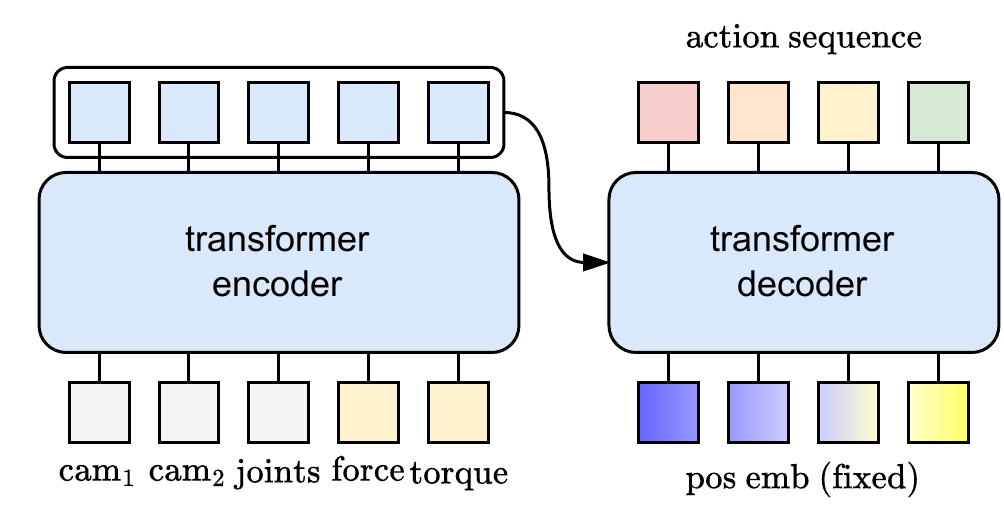}
    \caption{Model architecture: the transformer encoder is fed with the image from the gripper camera, the image of the head camera, the joints states, and the force and torque feedback. The remaining of the architecture is identical to \ac{act}.}
    \label{fig:architecture_diagram}
    \vspace{-2ex}
\end{figure}

\subsection{Dataset}
Human operators teleoperate the robot using a VR headset~\cite{vrheadset} remotely to collect the datasets. Each episode lasts approximately 2 minutes, during which the operators conduct the \ac{pnr}. In total, 412 episodes were collected. The image resolution is $480\times640$ for both the head and gripper images. The head view is formed by stitching the two head-camera images into a single panoramic frame, enabling effective manipulation in VR for operators. For training, the stitched panorama is consumed by the model as a single combined image. The dataset spans more than 10 kinds of bottles. One of the four bottles is lying on the desk, and the other three are standing upright.

\begin{figure*}
    \centering
    \vspace{10pt}
    \includegraphics[width=\linewidth]{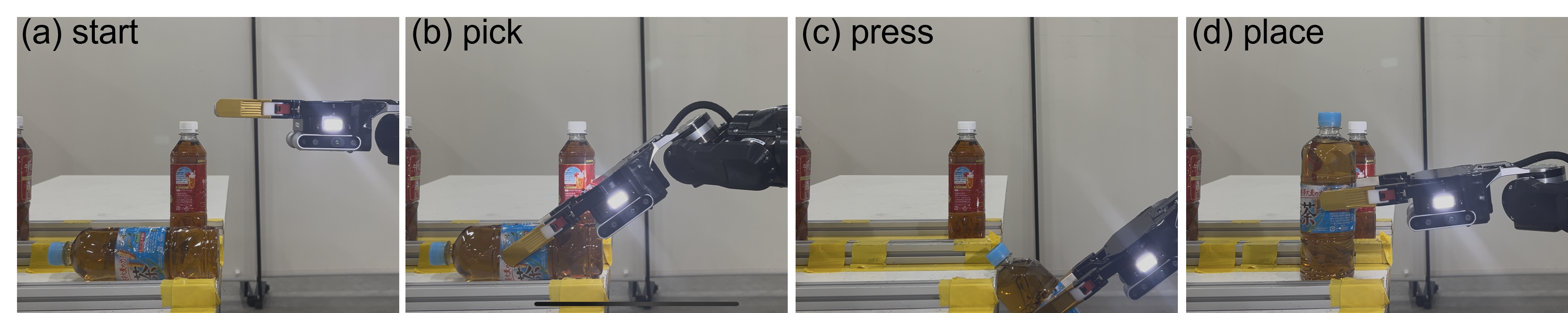}
    \caption{Bottle-recovery task decomposed into four stages: (a) \emph{Start}—initial state, (b) \emph{Pick}—approach and grasp the bottle, (c) \emph{Press}—move to the table edge and press the bottle against the it, (d) \emph{Place}—move to the table and set the bottle upright.}
    \label{fig:recovering_bottle}
    \vspace{-2ex}
\end{figure*}

\subsection{Training}
We use \ac{act} as implemented in \texttt{lerobot}~\cite{lerobot} as a baseline. Most hyperparameters follow the open-source configuration; we set the batch size to 96, the action-chunk size to 50, and train for 300{,}000 steps. All camera inputs are resized to $480\times640$ for the vision encoder. All 10 joint states are provided to the model, and the force and torque signals are concatenated with the joint states and fed directly into the model as shown in \figref{fig:architecture_diagram}.

\subsection{Inference}
During inference, the control loop runs at $50\,\mathrm{Hz}$ and outputs absolute joint position commands. The sensors' update rates are: head camera $10\,\mathrm{Hz}$, gripper camera $30\,\mathrm{Hz}$, joint states $50\,\mathrm{Hz}$, and force and torque $40\,\mathrm{Hz}$. These frequencies match those used during data collection for training.

%% file: sections/3-result.tex
\section{Experiments and Results}

\begin{figure}[!t]
    \centering
    \includegraphics[width=\linewidth]{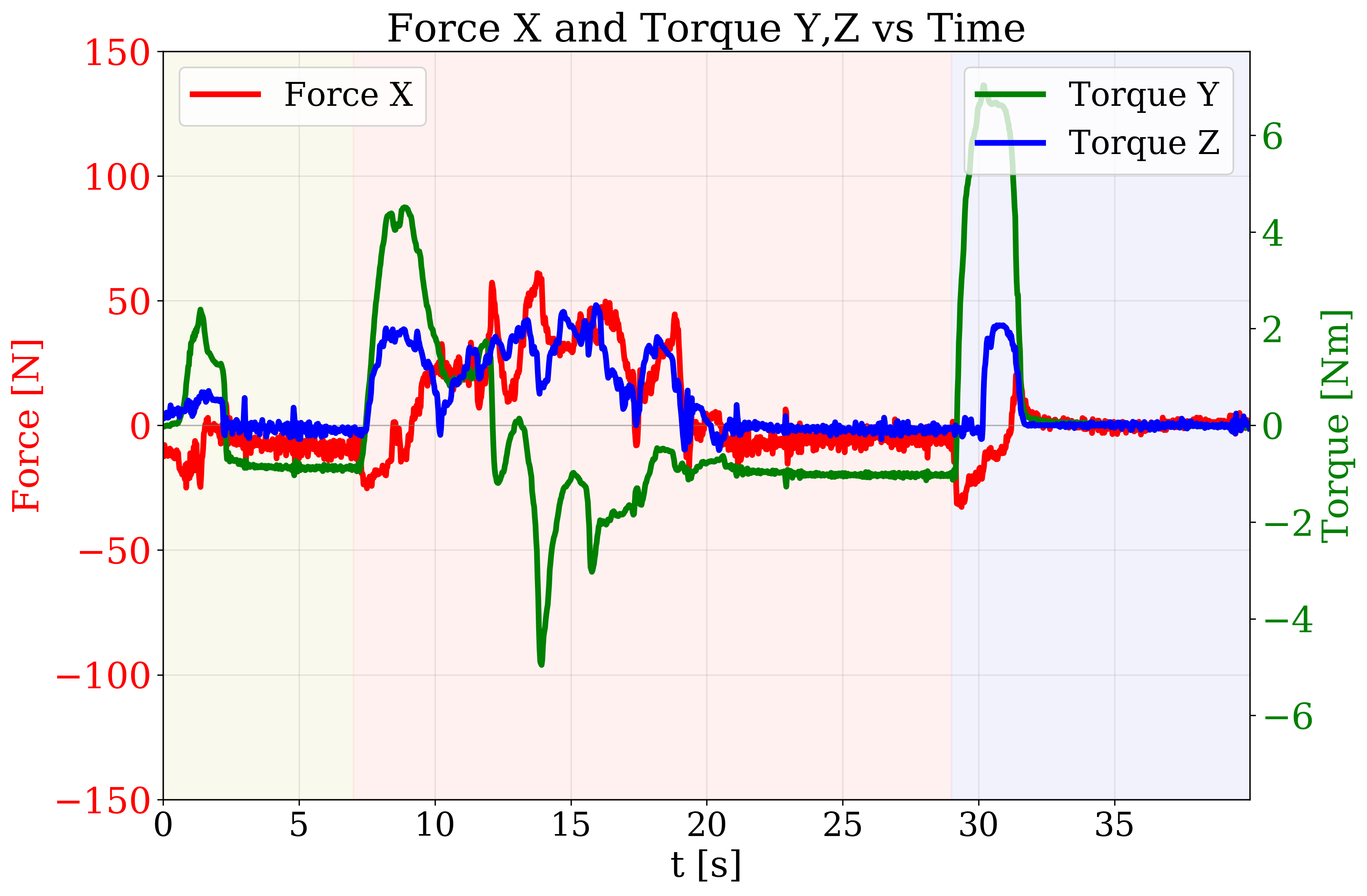}
    \caption{Time series of wrist force along the x-axis (red) and torques about the y- and z-axes (green, blue) during a bottle-recovery episode which have the most changes during the \ac{pnr} task. Shaded bands as a background color denote task phases: Pick (light yellow), Press (light red), Place (light blue). Large transients around 8–20 s and 29-32 s indicate contact events.}
    \label{fig:torque_force_analysis}
    \vspace{-2ex}
\end{figure}

\subsection{Bottle-Recovery Task}
A representative successful rollout of \ac{ftact} on the \ac{pnr} task is shown in \figref{fig:recovering_bottle}. Starting from an initial pose, the robot autonomously localizes and approaches the fallen bottle, grasps it with the gripper, and—mimicking the teleoperator strategy—presses the upper portion of the bottle against the table to reorient it. Finally, the robot places the bottle upright on the table. While bimanual systems~\cite{fu2024mobile_aloha, aldaco2024_aloha, raut2020learning} can accomplish this with two hands, it is challenging for a single-arm, gripper-equipped manipulator. We therefore deliberately exploit contact with the table to achieve reorientation, as teleoperators do.

\begin{table}[t]
  \centering
  \caption{Ablation on the Pick-and-Reorient task (Cumulative success rate \%)}
  \label{tab:ablation}
  \renewcommand{\arraystretch}{1.1}
  \setlength{\tabcolsep}{6pt}
  \begin{tabular}{llcccc}
    \toprule
    \multirow{2}{*}{Method} & \multirow{2}{*}{Objects} & \multicolumn{4}{c}{Pick-and-Reorient} \\
    \cmidrule(lr){3-6}
    & & Pick & Press & Place & Total \\
    \midrule
    \multirow{2}{*}{\shortstack{FTACT\\(ours)}}
      & Trained   & 100 & 100 & 100 & 100 \\
      & Untrained & 100 & 100 &  80 &  80 \\
    \midrule
    \multirow{2}{*}{\shortstack{ACT\\(baseline)}}
      & Trained   & 100 &  80 & 80 &  80 \\
      & Untrained & 100 &  80 &  60 &  60 \\
    \bottomrule
  \end{tabular}
\end{table}

\subsection{Ablation Study}
We evaluate the contribution of force and torque sensing with an ablation study. We compare two policies: (i) the force and torque augmented architecture described in the \emph{System} section (ours) and (ii) a baseline matching the observation space of \ac{act} which means that we excluded the force and torque as an observation. We consider two experimental settings— \emph{trained} and \emph{untrained} target objects for bottle types and spatial arrangements relative to the training data. Success rates are computed over five trials per scenario. 

As shown in \tabref{tab:ablation}, the proposed policy attains a higher total success rate than the variant without force and torque inputs. Consistent with the task decomposition in \figref{fig:recovering_bottle}, the gains are concentrated in the \emph{press} and \emph{place} stages, which are more contact-rich and demand finer manipulation.

\subsection{Analysis of Force and Torque}
To investigate the ablation results, we analyze the force-torque signals in \figref{fig:torque_force_analysis}. During the pressing phase (approximately 8–20\,s; red-shaded background), pronounced transients appear in torques about the \(y\)- and \(z\)-axes and in force along the \(x\) axis, corresponding to contact with the table. A second burst of activity occurs later in the episode (around 29–32\,s), associated with the placement and release the bottle event. These contact-driven signatures provide reliable cues that the bottle is being pressed against and subsequently placed on the table. Vision alone was not always sufficient to robustly detect these events for fine manipulation; incorporating force and torque feedback enabled the single-arm with gripper system to place the bottle more precisely.

%% file: sections/4-conclusion.tex
\section{CONCLUSION}
In this work, we introduced \ac{ftact}, an \ac{act}-based multimodal \ac{il} policy that augments images and joint states with wrist force and torque sensing for the contact-rich bottle \ac{pnr} task. In real-world robot evaluations, \ac{ftact} achieved greater overall success than a pure ACT baseline, especially in the press and place stages, where visual observability is limited. Signal analysis further showed consistent force and torque transients aligned with contact events, explaining why force and torque cues help disambiguate the task phase and improve precision during reorientation and placement. These results suggest a practical path to scaling retail manipulation: lightweight sensing combined with modern IL architectures can reduce human teleoperation without resorting to larger and slower models. 

While promising, our study is limited to one task family, a modest number of object and scene variants, and demonstration-driven learning. Future work will expand task diversity and evaluation scale, compare against diffusion policies and \ac{vla} models under matched compute, and explore policy refinement via online adaptation and safety-aware contact control. We also plan to investigate minimal sensing configurations and broader deployment considerations for store-scale autonomy.

%% file: acronyms.tex
\begin{acronym}
\acro{act}[ACT]{Action Chunking Transformer}
\acro{va}[VA]{Vision Action}
\acro{vla}[VLA]{Vision Language Action}
\acro{pnp}[PnP]{Pick-and-Place}
\acro{pnr}[PnR]{Pick-and-Reorient}
\acro{dof}[DoF]{Degrees of Freedom}
\acro{il}[IL]{Imitation Learning}
\acro{ftact}[FTACT]{Force and Torque aware Action Chunking Transformer}
\acro{ft}[F/T]{Force Torque}
\end{acronym}